\documentclass[runningheads]{llncs}
\usepackage{graphicx}
\usepackage{amsmath}

\begin{document}
\title{Learning the Designer's Preferences to Drive Evolution}

%
%

\author{Alberto Alvarez\inst{} \and
Jose Font\inst{}}
%
\authorrunning{A. Alvarez \and J. Font}
%
%
\institute{Malm\"{o} University, Nordenskiöldsgatan 1, 21119, Malm\"{o}, Sweden
\email{\{alberto.alvarez,jose.font\}@mau.se}}

\maketitle              
\begin{abstract}
This paper presents the Designer Preference Model, a data-driven solution that pursues to learn from user generated data in a Quality-Diversity Mixed-Initiative Co-Creativity (QD MI-CC) tool, with the aims of modelling the user's design style to better assess the tool's procedurally generated content with respect to that user's preferences. Through this approach, we aim for increasing the user's agency over the generated content in a way that neither stalls the user-tool reciprocal stimuli loop nor fatigues the user with periodical suggestion handpicking. We describe the details of this novel solution, as well as its implementation in the MI-CC tool the Evolutionary Dungeon Designer. We present and discuss our findings out of the initial tests carried out, spotting the open challenges for this combined line of research that integrates MI-CC with Procedural Content Generation through Machine Learning.

\end{abstract}

\section{Introduction}

As game production grows, so does the usage of computer-aided design (CAD) tools to develop various facets of games. CAD tools enable users to create new content or refine previously created content with the assistance of some type of technology that focuses on reducing the workload of the developer. Procedural Content Generation (PCG) denotes the use of algorithms to generate different types of game content, such as levels, narrative, visuals, or even game rules, with limited human input \cite{shaker_procedural_2016}. Search-based PCG is the subset of techniques whose approach generates content by using a search algorithm, a content representation mechanism, and a set of evaluation functions to drive the content creation process towards near-optimal solutions \cite{Yannakakis2018}. 

Mixed-initiative co-creativity (MI-CC)~\cite{yannakakis2014micc} is a branch of PCG through which a computer and a human user create content by engaging into an iterative reciprocal stimuli loop~\cite{shaker2013ropossum,smith_tanagra:_2011,machado2019pitako,liapis_generating_2013,guzdial-lvldsg-aiide-2018,lucas-3buddy-iccc2017}. This approach addresses the design process with insight and understanding of the affordances and constraints of the human process for creating and designing games \cite{Liapis2016}. MI-CC helps designers to either optimize their current design towards a specific goal (thus exploiting the search space) or foster their creativity by proposing unexpected suggestions (exploring the search space). To these ends, diversity has been an important feature for the research community to focus on during the past decade, including novelty search~\cite{Novelty-Lehman2011}, surprise~\cite{Surprise-Gravina2016}, curiosity~\cite{CuriositySearch-Stanton} and, more recently, quality-diversity approaches \cite{Khalifa2018}. 

PCG through Quality-Diversity (PCG-QD) \cite{gravina2019procedural} is a subset of search-based PCG, which uses quality-diversity algorithms~\cite{Pugh2016} to explore the search space and produce high quality and diverse suggestions. MAP-Elites \cite{Mouret2015} is a successful quality-diversity algorithm that maintains a map of good suggestions distributed along several feature dimensions. A constrained MAP-Elites implementation was presented by Khalifa et al.~\cite{Khalifa2018}, combining MAP-Elites with a feasible-infeasible (FI2Pop) genetic algorithm~\cite{Kimbrough2008} for the procedural generation of levels for bullet hell games. The first implementation of a PCG-QD algorithm for MI-CC was presented by Alvarez et al. \cite{alvarez2019empowering}, elaborating on the combined MAP-Elites and FI2Pop approach by introducing a continuous evolution process that benefits from the multidimensional discretization of the search space performed in MAP-Elites.

In all the above MI-CC approaches, the designers play an active role in the procedurally generated content while struggling between the expressiveness of the automatic generation and the control that they want to exert over it \cite{Alvarez2018}. Having this as motivation, this paper takes the work in \cite{alvarez2019empowering} one step forward by adding an underlying interactive PCG via machine learning algorithm \cite{summerville2018procedural}, the Designer Preference Model, that models the user's design style, to be able to predict future designer's choices and thus, driving the content generation with a combination of the designer's subjectivity and the search for quality-diverse content.



\section{Previous work}

\subsection{Mixed-Initiative Co-Creativity}
Similar to user or player modeling, designer modeling for content creation tools (CAD and MI-CC tools) was suggested by Liapis et al~\cite{Liapis2013-designerModel}, where it is proposed the use of designers models that capture their styles, preferences, goals, intentions, and interaction processes. In their work, they suggest methods, indications, and advice on how each part can be model to be integrated into a holistic designer model, and how each game facet can use and benefit from designer modeling. Moreover, in \cite{Liapis2014-designerModelImpl} the same authors discuss their implementation of designer modeling and the challenges of integrating all together in their MI-CC tool, Sentient Sketchbook, which had a positive outcome on the adaptation of the tool towards individual “artificial” users.

Furthermore, Lehman et al \cite{lehman2016creative} presented Innovation Engines that combine the capabilities and advantages of machine learning and evolutionary algorithms to produce novel 3D graphics with the use of Compositional Pattern-Producing Networks (CPPN) evolved with MAP-Elites, and evaluated by the confidence a deep neural network had on the models belonging to a specific object category.

\subsection{Procedural Content Generation via Machine Learning}
Summerville et al. \cite{summerville2018procedural} define Procedural Content Generation via Machine Learning (PCGML) as the generation of game content by models that have been trained on existing game content. The main approaches to PCGML are: autonomous content generation, content repair, content critique, data compression, and mixed-initiative design.

\begin{figure}
\includegraphics[width=\textwidth]{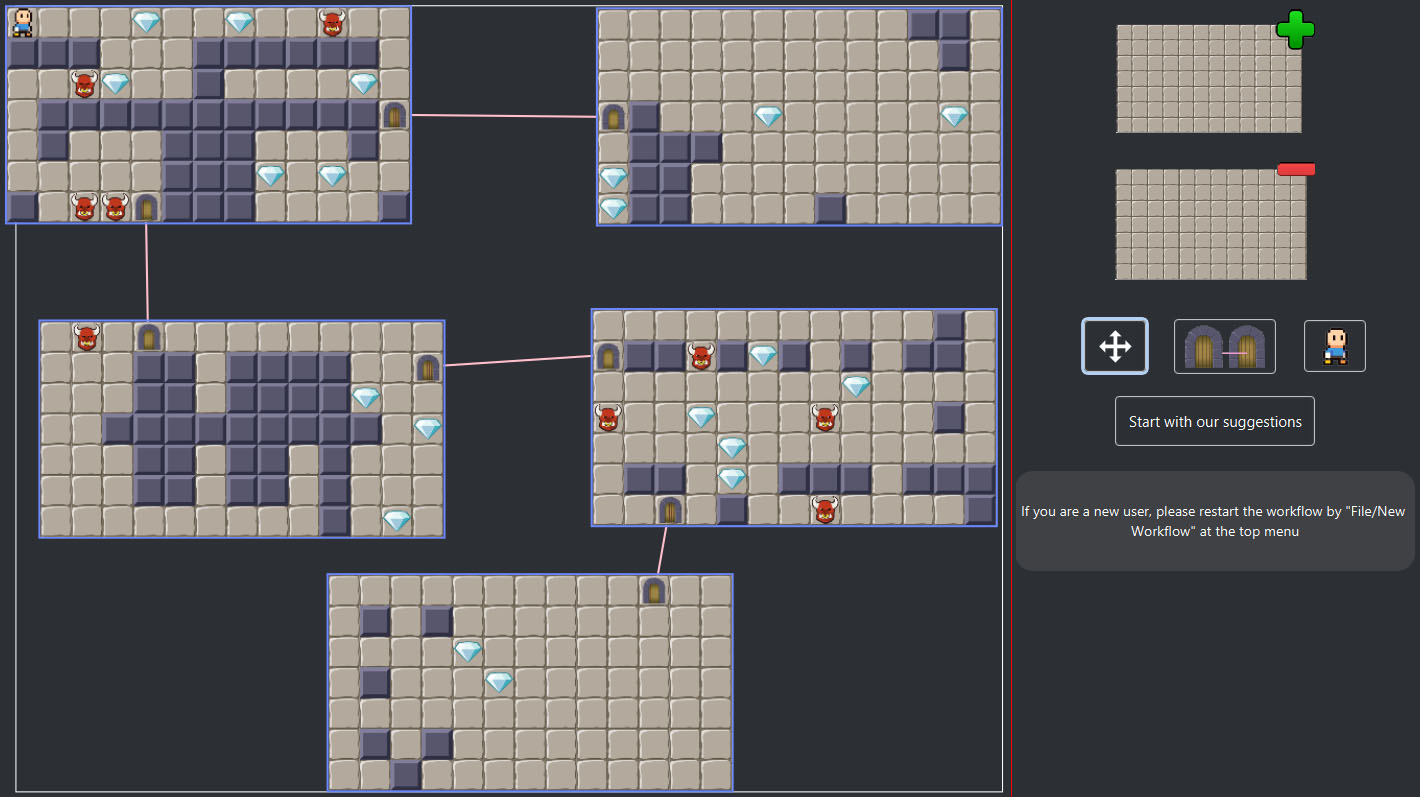}
\caption{Screenshot of the dungeon editor screen in EDD, displaying a sample dungeon composed by five rooms.} \label{fig1}
\end{figure}

In the latter case and, as appointed by Treanor et al. \cite{treanor2015ai}, AI may engage with a human user participating in the creation of content, so that new gameplay emerges from this shared construction. This emerging relationship between the user and the AI system, when implemented through a trained machine learning algorithm, has the potential to reduce user frustration, error, and training time. This is due to the capacity of a machine learning solution to adapt to the design preferences of the user that interacts with the MI-CC tool by learning from the user-generated dataset of previous choices.

\subsection{The Evolutionary Dungeon Designer}

\begin{figure}[t]
\includegraphics[width=\textwidth]{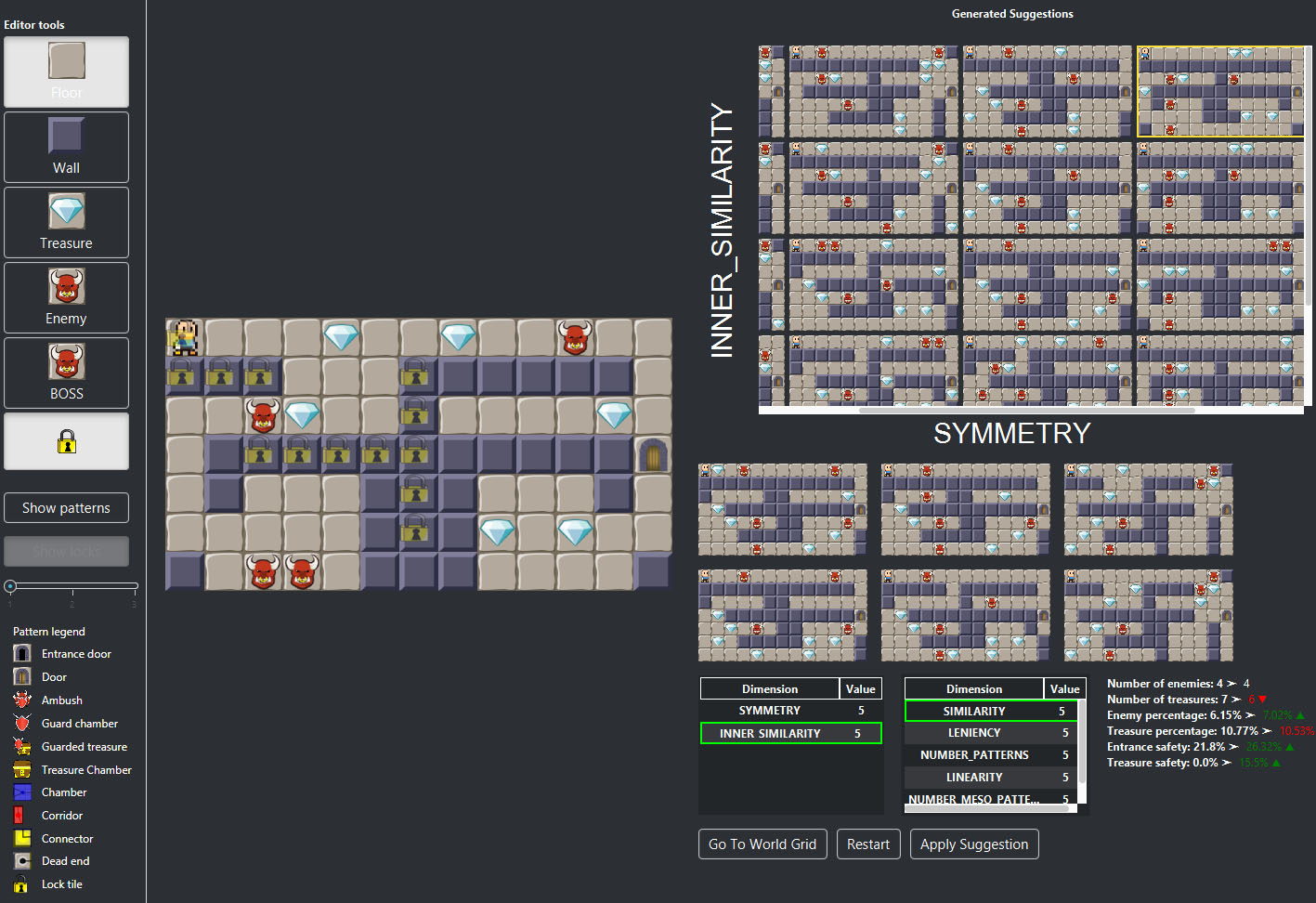}
\caption{The room editor screen in EDD. The top-right pane shows the suggestions provided by the IC MAP-Elites algorithm. Below are the six top-raked suggestions by the Designer Preference Model. The left pane contains the manual edition features.} \label{fig2}
\end{figure}

The Evolutionary Dungeon Designer (EDD) is an MI-CC tool for designers to build 2D dungeons. EDD allows designers to manually edit the overall dungeon and its composing rooms (see Figure \ref{fig1}), as well as to use procedurally generated suggestions either as inspiration to work on or as a finished design (see Figure \ref{fig2}). Both options fluently alternate during the creation process by means of a workflow of mutual inspiration, through which all manual editions performed by the user are fed into the underlying continuous Evolutionary Algorithm, accommodating them into the procedural suggestions. A detailed description of EDD and its features can be found in~\cite{Alvarez2018a,Alvarez2018,Baldwin2017a,Baldwin2017}.

Subsequent user studies \cite{Alvarez2018,Baldwin2017} carried out with game designers on EDD raised the following areas of improvement: (1) the designers struggled with EDD’s capability of understanding the designer’s intentions and preserving custom designs; (2) the tool was unable to generate aesthetically pleasing suggestions since the fitness function only accounted for functionality, but not aesthetics, of design patterns; (3) the designers wanted to keep certain manual editions from being altered by the procedural suggestions.  

With the aims of addressing these limitations as well as fostering the user's creativity with quality-diverse proposals, EDD was improved with the Interactive Constrained MAP-Elites (IC MAP-Elites) \cite{alvarez2019empowering}, an implementation of MAP-Elites into the continuous evolutionary process in EDD. With this addition, the user drives the generation of procedural suggestions by modifying at any moment the areas of the search space where the evolution should put the focus on. This is done by selecting among the available dimensions: symmetry, similarity, design patterns, linearity, and leniency. Additionally, the designers have now the chance to limit the search space by locking map areas and thus preserving manually edited content.

This paper contributes by building on top of EDD's IC MAP-Elites, adding a data-driven Designer Preference Model that adapts and personalizes the design experience, as well as balances the expressivity of the tool and the controllability of the designer over the tool. Other researchers have pursued a similar goal by biasing the search space through having the user perform a manual selection after every given number of generations~\cite{Picbreeder-Secretan2008,Liapis2012-adaptiveVisual,Novelty-Lehman2011}. Nevertheless, this approach leads to an increase in user fatigue by repeatedly asking for user input and thus, stalling the evolutionary process until such input is received. Moreover, this staged process seems incompatible with the dynamic reciprocal workflow of MI-CC tools, where the focus is on the designer proactively creating content rather than passively browsing a set of suggestions.

The remaining sections of the paper are structured as follows: Section 3 describes the data-driven Designer Preference Model; Section 4 presents the initial experimental results, and Section 5 discusses the results and future lines of research of this novel approach.

\section{Designer Preference Model} \label{section/model}

\begin{figure}
\includegraphics[width=\textwidth]{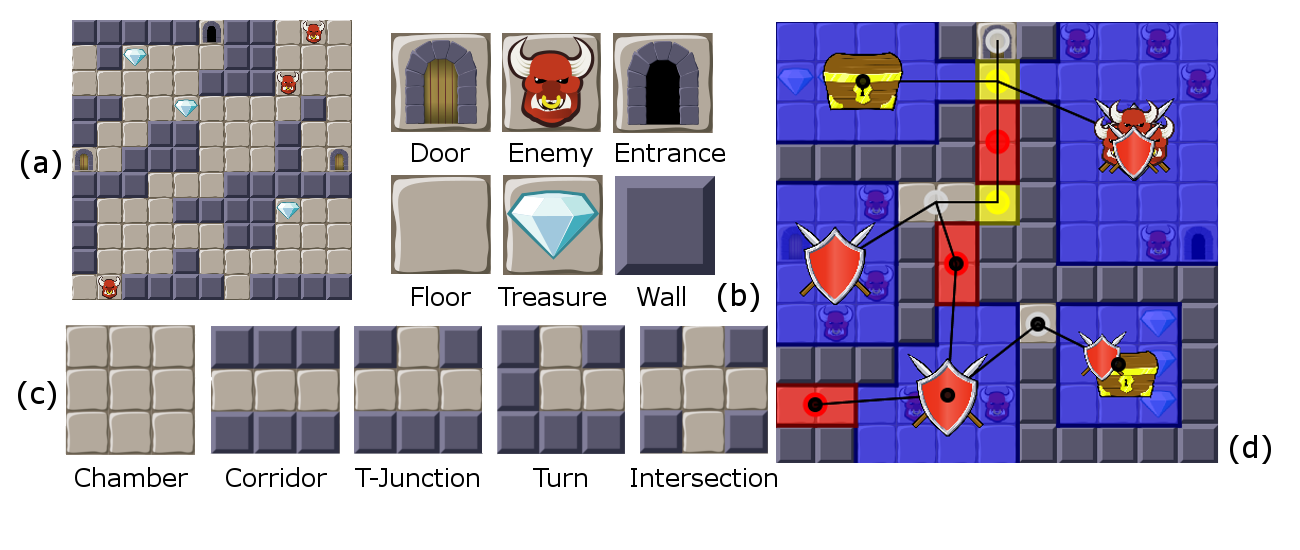}
\caption{A sample room in EDD (a) compose by tiles (b), spatial patterns (c) and meso-patterns (d). Detailed descriptions for these components can be found in~\cite{Baldwin2017a}.} \label{fig3}
\end{figure}

The Designer Preference Model is a data-driven intelligent system that learns the user's design style by training and testing over a continuously growing dataset composed of the user's actions and choices while operating EDD. The underlying evolutionary algorithm (EA) uses this model to assess the generated suggestions according to the predicted preference of the designer. This is a complementary assessment to EDD's original fitness function, which evaluates individuals first based on the presence and distribution of spatial and meso-patterns (Figure \ref{fig3}), and then based on their degree of adaptation to the user-selected quality-diversity dimensions \cite{alvarez2019empowering}. The relevance of the Designer Preference Model gradually increases over EDD's fitness function as long as the model gains confidence in its assessments.

\begin{figure}
\includegraphics[width=\textwidth]{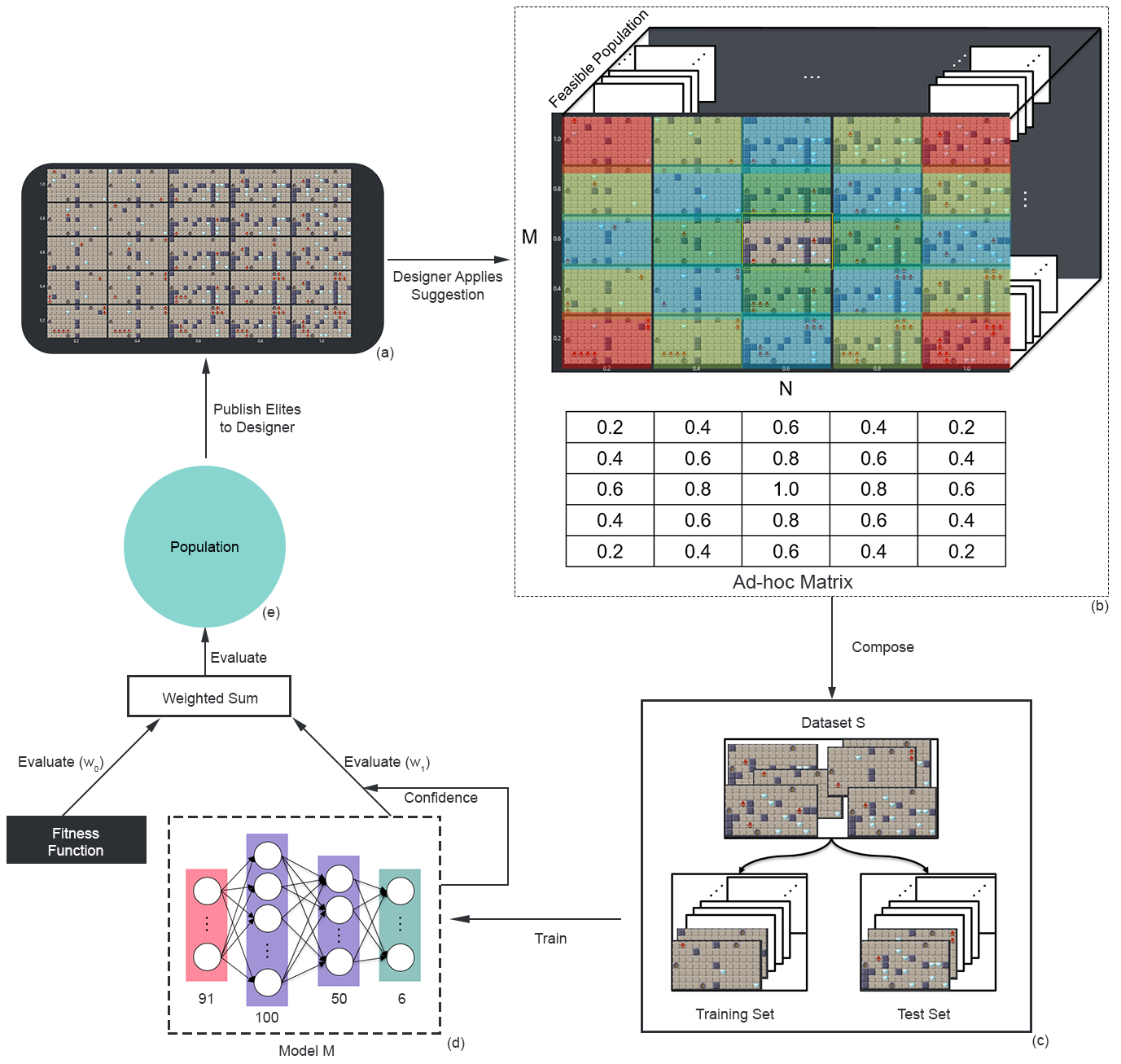}
\caption{Overview of the Designer Preference Model integrated into the fitness function of EDD. Elites are published and shown to the designer in a grid fashion (a), and once the designer chooses and applies one of the suggestions, an ad-hoc matrix is created based on the position of the selected suggestion to estimate the preference of suggestions (b). The ad-hoc matrix is then applied to all the elites in the grid, and the feasible populations within the EA cells to compose a general dataset $S$ with rooms labeled by the estimated preference. The composed dataset $S$ is then subdivided into a training set (90\%) and test set (10\%), both with the same label distribution (c). The dataset is used to train a model $M$, which is a relatively small neural network, for 20 epochs (d). The  model is then used to evaluate the population of the EA together with the current fitness function in a weighted sum, with the weight of the model $M$ conditioned by the confidence of the network (e).} \label{fig4}
\end{figure}

\subsection{Model Update and Usage}
The proposed model is a relatively small neural network $M$ with as many input neurons as the number of tiles composing each room, two hidden layers (100 and 50 neurons respectively), and six output neurons, one per each discrete preference value assigned to the individuals by the designer. When the designer starts EDD, the neural network is created with random initialization and without any prior training (i.e. cold start). While the designer creates and modifies rooms, on the background, the EA produces and presents individuals to the designer using the MAP-Elite’s cells (Figure \ref{fig2}), while it adapts to the designer’s design. Following a proactive learning approach \cite{donmez2008proactive}, anytime the designer chooses one suggestion to replace her current design, a training session is requested for a model $M$ with a dataset $S$ created with the current cells and their populations based on the designer’s chosen suggestion. The loop, depicted in figure \ref{fig4}, can be described in the following two steps:

\subsubsection{Dataset creation:}

The designer chooses a suggestion to replace her current design, which in turn, requests a training session using all the current individuals (i.e. the elites and the rest of the feasible populations) to create a new dataset to train the model closer to the “actual” preference of the designer. As shown in figure \ref{fig4}.b, an ad-hoc matrix is created, based on the position of the applied suggestion, to calculate the estimated preference, starting with the applied suggestion (1.0 preference value), and reducing the preference value by 0.2 per each step that was taken away from the applied suggestion in the matrix until a minimum of 0.0. 

Once all the individuals are given an estimated preference value based on their grid position by the ad-hoc matrix, they are all used to compose a general dataset $S$ where each individual is transformed to match the network input. Finally, we divide the set into a training set (90\%) and test set (10\%) with the same label distribution. Through this process, we end up having a maximum of $M \times N \times feasible_{population}$ tuples, which relates to the granularity of each presented dimension times the maximum amount of feasible individuals per cell.

\subsubsection{Training and usage:}

The model is then trained for a limited set of epochs (i.e. 20 epochs) and later incorporated into the evolutionary loop to further evaluate individuals. As mentioned above, the model tries to slowly fit towards the designer’s preference, and as it becomes more confident in predictions, the more weight $W_{1}$ it has in the final fitness of an individual. Confidence is calculated based on the output of the softmax layer, which 
limits the output of all the neurons into the range 0 to 1, as the sum of all the neurons' output must be 1.0. This characteristic of the softmax layer enables us to interpret the results as the probabilities for each of the classes. For instance, if the network predicts that an individual is going to be preferred to the designer with a 1.0 preference with a probability of 0.9, it means that the remaining 0.1 is distributed among the other output classes, and as a consequence, the network has high confidence. The resulting weights (Eq. \ref{eq:weights}) and weighted sum (Eq. \ref{eq:weightedSum}) to evaluate each of the individuals in the EA were the following:


\begin{equation} \label{eq:weights}
\begin{split}
 w_{1}={}&\min(M_{conf} \cdot M_{TestAcc}, 0.5),\\
w_{0} ={}& 1.0 - w_{1}   
\end{split}
\end{equation}


\begin{equation} \label{eq:weightedSum}
weightedSum = (w_{0} \cdot objective) + (w_{0} \cdot predicted_{pref})
\end{equation}

Finally, the loop continues and the model awaits for the next training session that will be triggered the next time that the user applies a suggestion. In the meantime, the trained model is used as part of the combined individual evaluation process.
\section{Evaluation} \label{section/experiment}


\subsection{Model performance, integration, and setup}
We conducted a set of experiments to test the extent to which the Designer Preference Model learns from the user-generated data and fits into the previously existing MI-CC workflow in EDD. These experiments also aimed for finding the hyperparameter configuration for the model that better suited its goals. 

This resulted in a fully connected neural network with two hidden layers with 100 and 50 neurons respectively. Bigger and deeper networks, as well as longer training epochs, did result in higher accuracy but it was not worth the time-complexity/accuracy tradeoff since it obstructed the dynamic and high-paced workflow of the tool. Finally, the network had six output nodes related to the different preference values a suggestion could have (i.e. from 0.0 to 1.0 in 0.2 intervals, both ends inclusive) with a softmax layer, which was used to account for the confidence on the network.

Additionally, we decided to train the model's network under independent episodes every time the designer applied a suggestion using the most up-to-date data (the dataset that was created each time a selection was applied). We evaluated and through experimentation later discarded a more continuous approach, since continuously training between episodes led to the generation of large noisy datasets that distorted the training process. 

As a result, the Designer Preference Model is smoothly integrated into EDD's workflow. User-wise, it runs in a completely transparent way, neither breaking the reciprocal stimuli loop nor slowing down the performance of the EA in a perceptible way. 

\subsection{User Study}
A user study was also conducted to collect preliminary results that assess the relevance of the Designer Preference Model. We aimed for gathering feedback from game designers on how the model would be used, as well as their perception of the adaptive capabilities of the model. 

Fifteen game design students (i.e. novice designers) participated in the study; all of them were introduced to all the features of the tool and were tasked to create a dungeon with interconnected rooms for as long as they were satisfied with their design. At the end of each test session, the participants were asked to fill a brief questionnaire assessing their understanding of the suggestions, its usability, pros, and constraints. 

For the purposes of the user study and to test the new model's assessment capabilities in contrast to EDD's original fitness function, we presented the suggestions as displayed in Figure \ref{fig2}. The top-right pane displays EDD's IC-MAP-Elites as described in~\cite{alvarez2019empowering}. The bottom-right pane shows a smaller grid displaying the top raked individuals assessed by the Designer Preference Model. As the designer applied the top suggestions, the lower grid would get trained with the expected preference, as explained in section \ref{section/model} and, as a consequence, the lower grid would become more adapted.

This system was designed to validate the hypothesis that users would prefer to make use of the suggestions in the bottom-right pane in the long run, after the Designer Preference Model had been trained a sufficient amount of times, thus gaining confidence in its assessment. A total of 105 rooms were created and the designers applied 43 times suggestions to their designs, with most of the cases happening once the designers had manually created most of the dungeon. Unfortunately, this did not generate enough activity in EDD's procedural content generation system to be able to draw accurate conclusions from the study.   

\section{Open Problems and Future Work}


 
This paper presents the first MI-CC tool with quality-diversity that explores the usage of a data-driven designer preference model, and its implementation into the EA loop as a complementary evaluation of individuals. Through this model, we searched to cope with some of the limitations presented in previous work, mainly, the user fatigue when queried to choose solutions for the EA, and the stalling of the evolutionary process, thus, adapting the control of the user in the search-space to the dynamic workflow of MI-CC tools. 

In this section, we present the multiple challenges that arose when trying to use the designer preference model from our first experiments and preliminary study and the open areas for active research. Through our user study, we were able to test the behavior of our preference model adapted to each of the designers and the performance of such in the wild. While the model, in general, was less used than expected, it was indeed able to learn to certain extent characteristics of the preferred suggestions. 

\subsection{Dataset}

The dataset $S$ created each discrete step the designer applied a suggestion, had a set of intrinsic attributes that while positive and interesting to learn from, they could have been counterproductive and could potentially explain the low and fluctuating accuracy of the model. Firstly, as mentioned in section \ref{section/model}, each generated dataset had a maximum number of samples of $M \times N \times feasible_{population}$, capped to 625 samples in our study, which might not be enough data to accurately learn or would require more training epochs, which ultimately would result in overfitting. This aligns with the open problems presented in~\cite{summerville2018procedural}, where the authors discuss that games will always be constrained by the amount of data, and even though we can generate many samples with our EA, it still might not be enough to cope with the amount of data that ML-approaches require.

Secondly, by taking advantage of the grid visualization of the MAP-Elites, we also inherited the behavioral relation among the different elites, and consequently, each independent training session would intrinsically represent such relation. While our objective was indeed to learn this behavior relationship, which could reveal interesting relations and perspectives by the model, the differences that each pair of behavioral dimensions have could potentially disrupt the whole model between training sessions. For instance, if we train with symmetry and similarity as dimensions, and subsequently change them to symmetry and leniency, what before could be 0.8 in preference in the dataset (i.e. a neighbor of the previously applied suggestion), could now be 0.0 in preference for this dataset, since the pair of dimensions would sort individuals completely different.

Finally, the fact that we automatically assigned an estimated preference value to all individuals based on their grid position, and as pointed out in the previous point, relations could fluctuate dramatically, which could arise a potential issue with the dataset. For instance, a challenge with estimating the preference can be observed in the aesthetic aspects of the rooms, where two rooms can be quite aesthetically similar (i.e. have a single different tile) and yet, due to the way we assign the preference values to train, have a very different preference, thus, enabling confusion in the model. Nevertheless, we did not want the assigned preference value to be based on the similarity between suggested rooms since what the model would end up just learning is to classify based on aesthetic similarity. Therefore, there would not be any need to train any model and through just composing a similarity table and comparing new rooms to the ones already included we would probably achieve the same result.


\subsection{Preference modality}

We chose the suggestion grid of the MAP-Elites as an inflection point for the training of the model since it felt more appropriate and natural to the workflow of the tool, and more of a pointer to the actual preference of a designer. The suggestion grid is a reflection of the EA search for quality solutions and having the designer proactively choosing solutions that were interesting for them seemed like an indicator of the preference and interest of the designer.

Based on when the designers actually started applying suggestions and their reason why, indicates that they were not as representative of the preference of the designer as expected. Instead, suggestions were seen as an in-between step to help shape the final room, after creating a first draft of the room and before actually reaching a satisfactory room. This opens up the investigation on what design processes or combinations of processes could be captured to accurately represent the designers’ preferences with higher fidelity. 

Firstly, we need to consider the level of the designer that is using the tool. The design process, the objectives when designing, the vision on what to do, and the ideas on what to design and what is expected from an interactive tool as ours, could vary quite drastically between designer levels, as it is concluded in\cite{lucas-3buddy-iccc2017}. Considering our previous studies with game designers that are more experienced and the one done for this study, we realize that novice designers come with many different ideas that they would like to try, as well as experimenting with very different designs, which in turn means that their preferences and intentions change in very short periods. Understanding this, and adding it as a constraint on the design of preference models is vital since we would want to recognize this key changes to probably discard the model and start fresh since what the model had learned might not be useful anymore.

Secondly, choosing what and when to gather information to create the model is a key aspect. Besides the EA suggestions on the designer’s design, we could use the designer’s history of changes through their design as well as their current designs. In our case, constantly analyzing the composition of the dungeon and the rooms could bring some insight on the stage of the design process of the designer, which could be used to further understand what to use, if we should keep using the same model, and how to train. 

It might even be relevant to have a set of models per set of rooms that have some qualitative similarities to avoid confusion in the model, and updating the model that is relevant to the specific objectives of the designer. In counterpart, this would break the aspect of generalization (i.e. learning the preference of the designer throughout their design process) that could enable us to learn more from the designer.




\subsection{Dynamic-Dynamic System vs. Dynamic-Static System}

In our experiments, we designed a system where the model would move through the solution space (i.e. the preference-space of the designer) as the designer moves as well, which we call a dynamic-dynamic system. In such a system, the designers drift in many dimensions as they develop, understand better the tool, get deeper in the creative process, have different objectives, and such on. Further, designers might have drifted quite drastically in between training sessions, which ultimately makes the dynamic model harder to move with the designers, resulting in a deficient model. 

Therefore, we can conclude that to have some stability and be more robust to an ever-changing designer and creative process, we need some part of the approach to be static. Yet, the designer will never stop being a dynamic component, thus, it is the model that needs to be static. An exciting and interesting open area of research is then in the notion of community models, which would be models fed with several designers’ designs, clustered together by their qualitative similarities creating archetypes of designers or archetypes of designs. Such a set of group models would adapt to the dynamic designer by placing the models in the solution space, where a designer instead of drifting together with their model, they would traverse such a space of models as she drifts through the many dimensions of her creative process.

\subsection{Future Work}

Taking as a starting point the big amount of data (i.e. handmade rooms) collected from all the user studies done to date, and as abovementioned, we believe that a community model formed through clustering is a more realistic model. The envisioned system would follow exactly the same approach and core concept presented in this paper, i.e. a model that as it becomes more confident on the preference of the designer, the more weight it has to evaluate newly generated individuals by the IC-MAP-Elites, as a complementary evaluation to the objective function. 

Such a system could be created by using the data of each designer (i.e. a list of created rooms), then those could be arranged in different clusters that would represent archetypical designers or archetypical designs. From this point, we would have a foundation from which we would categorize new designers and we could, on the one hand, create a model from the data in the cluster and start adapting it to the current designer, avoiding the cold start problem. On the other hand, we could as well just keep trying to assign the designer, based on her designs, to different clusters, using each cluster as a model to infer what the “community” of designers would prefer, and since, the designer is part of that community at the moment, what she would prefer. Therefore, creating a model that could be more robust for evaluating designers’ preferences by means of having more or less stable clusters that designers could navigate as they go deeper into the design process.

Furthermore, we could go a step further and conceptualize a layered model that on the top layer could represent the community models of the designers, and on the bottom layer, specific designer's models. The bottom layer would then be created in a more classical training session outside our MI-CC tool, with the designer being queried a set of models and she explicitly labeling what she likes and whatnot.
Such a model could be used to communicate the expected design style and preference among a group of designers working together or to train new designers based on senior designers' preferences, intentions, and style. 

We would also like to explore different steps on the tool where we could collect relevant and crucial data of the designers that could bring us a step closer to a more accurate model of their preferences. Furthermore, accounting for the designer level could have a very impactful result on an effective model, and on how we handle them and their relevance.

Finally, exploring and using different representations of the data, such as images of the rooms in a Convolutional Neural Network (CNN), or qualitative and more processed information of the room (e.g. tiles density, sparsity, and amount, room complexity, connected rooms information, etc.) is an interesting future line. We believe that CNNs could perform better but required even larger amounts of data, and creating 625 images of the suggestion (i.e. our maximum number of data tuples) and then training the model could be cumbersome and have a significant impact on the workflow.

\section{Conclusion}

In this paper, we have presented the Designer Preference Model, which is a data-driven system that learns an individual designer's preference through the designer’s proactive choosing of generated suggestions without disrupting the continuous reciprocal workflow in MI-CC. We implemented our approach in the Evolutionary Dungeon Designer, a Quality-Diversity MI-CC tool, where designers can create dungeons and rooms while the underlying evolutionary system provides suggestions adapted to their current design. 

We used the model as a complementary evaluation system to the fitness function of the suggestions in a weighted sum, where the model gained more weight as it became more confident and performed better. Therefore, we aimed at better assessing these provided suggestions with the use of the Designer Preference Model, for them to be interesting and preferable but still usable for designers.

Through our experiments and preliminary studies on using the model to adapt to different designers, we identified a set of challenges and open areas for active research that integrates MI-CC with PCG through Machine Learning. Those challenges relate to the amount of user data needed to accurately learn from the user's preferences, what type of data is needed from the process, the cold start problem, the seldom collection of data to train, the quality of the dataset, and the designer-model setup. Moreover, we wanted to come closer to machine teaching~\cite{simard2017machineTeaching} approaches where the human provides fewer data points but with higher quality (i.e. the necessary data to correctly learn) rather than classic approaches to ML (i.e. offline training with a substantial amount of data). In our approach, while the designer has the decision on when to train the algorithm and to a certain extent, with what data to train, we are still missing certain granularity to empower designers to give the right information to the algorithm.

The combination of MI-CC tools with PCG through Machine Learning is a promising area of research that has the potential to enhance content creation. Specifically, designer modeling and our approach to model the designer's preference can have a great impact on the creative process of designers by considering their preferences, intentions, and objectives into the loop, by adapting the workflow to their requirements, or by smoothing the communication among various designers.

Finally, by adding the preference model as a complementary evaluation to the generated suggestions of the evolutionary algorithm, we can give more control, to a certain extent, to the designers over the evaluation of the individuals. In consequence, we can generate higher quality suggestions that better fit a specific designer.



\section*{Acknowledgements}
 The Evolutionary Dungeon Designer is part of the project \textit{The Evolutionary World Designer}, supported by The Crafoord Foundation.

\bibliographystyle{splncs04}
\bibliography{references.bib}
%




\end{document}